\UseRawInputEncoding
\documentclass[conference]{IEEEtran}
\usepackage{cite}
\usepackage{multirow}
\usepackage{amsmath,amssymb,amsfonts}
\usepackage{algorithmic}
\usepackage{graphicx}
\usepackage{textcomp}
\usepackage[dvipsnames]{xcolor}
\usepackage{url}
\usepackage{hyperref}
\usepackage[editing]{coop-writing}
\newcommand*{\crosssymbol}{%
    \text{%
      \raise 1ex\hbox{%
        \rlap{\vrule height.2pt depth.2pt width .75ex}%
        \hbox to .75ex{\hss\vrule height .5ex depth 1ex\hss}%
      }%
    }%
}

\usepackage{xparse}
\usepackage{listings}
\usepackage{xcolor}

\definecolor{eclipseStrings}{RGB}{42,0.0,255}
\definecolor{eclipseKeywords}{RGB}{127,0,85}
\colorlet{numb}{magenta!60!black}

\lstdefinelanguage{json}{
    basicstyle=\normalfont\ttfamily,
    commentstyle=\color{eclipseStrings}, 
    stringstyle=\color{purple}\ttfamily,
    numbers=left,
    numberstyle=\scriptsize,
    stepnumber=1,
    numbersep=8pt,
    showstringspaces=false,
    breaklines=true,
    frame=lines,
    string=[s]{"}{"},
    comment=[l]{:\ "},
    morecomment=[l]{:"},
    literate=
        *{0}{{{\color{blue}0}}}{1}
         {1}{{{\color{blue}1}}}{1}
         {2}{{{\color{blue}2}}}{1}
         {3}{{{\color{blue}3}}}{1}
         {4}{{{\color{blue}4}}}{1}
         {5}{{{\color{blue}5}}}{1}
         {6}{{{\color{blue}6}}}{1}
         {7}{{{\color{blue}7}}}{1}
         {8}{{{\color{blue}8}}}{1}
         {9}{{{\color{blue}9}}}{1}
}
\def\BibTeX{{\rm B\kern-.05em{\sc i\kern-.025em b}\kern-.08em
    T\kern-.1667em\lower.7ex\hbox{E}\kern-.125emX}}
\begin{document}

\title{JSON-Bag: A generic game trajectory representation}

\author{\IEEEauthorblockN{Dien Nguyen, Diego Perez-Liebana, Simon Lucas}

\IEEEauthorblockA{\textit{Queen Mary University of London - Game AI Group} \\ 
{d.l.nguyen, diego.perez, simon.lucas}@qmul.ac.uk}
}

\maketitle
\IEEEpubidadjcol

\begin{abstract}
We introduce JSON Bag-of-Tokens model (JSON-Bag) as a method to generically represent game trajectories by tokenizing their JSON descriptions and apply Jensen-Shannon distance (JSD) as distance metric for them. Using a prototype-based nearest-neighbor search (P-NNS), we evaluate the validity of JSON-Bag with JSD on six tabletop games---\textit{7 Wonders}, \textit{Dominion}, \textit{Sea Salt and Paper}, \textit{Can't Stop}, \textit{Connect4}, \textit{Dots and boxes}---each over three game trajectory classification tasks: classifying the playing agents, game parameters, or game seeds that were used to generate the trajectories.

Our approach outperforms a baseline using hand-crafted features in the majority of tasks. Evaluating on N-shot classification suggests using JSON-Bag prototype to represent game trajectory classes is also sample efficient. Additionally, we demonstrate JSON-Bag ability for automatic feature extraction by treating tokens as individual features to be used in Random Forest to solve the tasks above, which significantly improves accuracy on underperforming tasks. Finally, we show that, across all six games, the JSD between JSON-Bag prototypes of agent classes highly correlates with the distances between agents' policies. 
\end{abstract}


\newcommand\width{0.75}
\section{Introduction}
Defining features and representations for games and their corresponding distance/similarity metric is foundational for any task that requires game analysis. Designing agents to play a game in a certain way (either to optimize playing strength \cite{silver_temporal-difference_2012}, model human players \cite{tychsen_defining_2008}, or optimize playstyle diversity \cite{guerrero-romero_map-elites_2021}) often requires hand-crafted features using domain knowledge. Automated game design and content generation requires defining game metrics to evaluate generated solutions \cite{browne_automatic_2008}. In these tasks, instead of only optimizing for the targeted fitness functions, optimizing also for diversity and novelty in the solution population can produce better results \cite{lehman_abandoning_2011} \cite{guerrero-romero_map-elites_2021}. Diversity in the population is usually enforced by either defining behavior criteria that partition the search space \cite{mouret_illuminating_2015} or using a distance metric to evaluate the novelty of new solutions \cite{lehman_abandoning_2011}. Data mining and analysis of game data, such as playstyle clustering \cite{drachen_comparison_2013}, also use distance and similarity metrics to discover patterns in the data.

In the majority of cases, features and representations are defined manually using domain knowledge or automatically using deep-learning models. The latter case often still requires feature engineering and further tuning to adapt to specific use cases. The distance and similarity metrics of choice are typically Euclidean distance \cite{drachen_comparison_2013} and cosine similarity \cite{zhan_retrieving_2018}.

This work proposes the JSON Bag-of-token model (JSON-Bag) to generically represent game trajectories by tokenizing their JSON descriptions. The JSON representation for a game trajectory is simply formed by concatenating the JSON of individual game states, serialized from the game objects' data, into a list. A JSON-Bag is a collection of token-occurrence count pairs from tokenizing the game trajectory in JSON. We interpret JSON-Bag as a probabilistic model of game trajectories and use the Jensen-Shannon distance (JSD) \cite{lin_divergence_1991} to measure similarity between them.

To validate our approach, we test it on six tabletop games---\textit{7 Wonders}, \textit{Dominion}, \textit{Sea Salt and Paper}, \textit{Can't Stop}, \textit{Connect4}, \textit{Dots and boxes}, each on three classification tasks: classifying different playing agents, game parameters, and game seeds. For each task, a classifier uses JSON-Bag as the representation to classify game trajectories into the correct class of agents, game parameters, or game seeds that have been used to generate the trajectories. Using a prototype-based nearest neighbor search algorithm (P-NNS), we compare JSON-Bag with JSD against a baseline with hand-crafted features and Euclidean distance and show that our approach outperforms the baseline in the majority of cases.

The first three games above have many unique game components, therefore sensible hand-crafted features are easy to define and the JSON-serialized game states are rich in information. We call the other three games ``sparse" games, meaning they have few unique game components. For example, \textit{7 Wonders} has coins, material resources, and multiple unique cards, while \textit{Connect4} only has a grid with a unique board piece type for each player. For sparse games, their JSON descriptions are therefore not as informative and P-NNS performance degrades. However, by applying Random Forest (RF) with minimal tuning, JSON-Bag performance significantly improves on tasks P-NNS underperformed, especially on sparse games. This further validates JSON-Bag's ability for automatic feature extraction and suggests the potential to use JSON-Bag with more sophisticated feature selection methods and models that can learn interactions between features.

A JSON-Bag prototype of a class is a single JSON-Bag that best represents a class of game trajectories. We evaluate N-shot classification for JSON-Bag P-NNS to show that JSON-Bag prototypes efficiently represent game trajectory classes, only needing a few samples per class to accurately classify classes with ``obvious" differences. This is useful for the application of diversity/novelty search, such as in playstyle modeling \cite{guerrero-romero_map-elites_2021}, to determine the novelty of new solutions. JSON-Bag prototype being sample efficient means a new solution would not require many game trajectories for its novelty to be estimated by JSD.

We compare the JSD between JSON-Bag prototypes of different playing agent classes to their actual behavioral differences. The JSD of JSON-Bag prototypes is found to be highly correlated with the average distance in the agents' policies.

The main contributions of this paper are:
\begin{itemize}
  \item \textbf{JSON-Bag}: A method to generically represent game trajectories using only the JSON descriptions of individual game states for tokenization. We describe a domain-agnostic method of tokenizing that works well across a diverse range of tabletop games. JSON serialization is general and many game implementations are able to either directly serialize game states to JSON or require little coding effort. JSON-Bag enables comparing JSON game logs without further domain-specific processing.
  \item We show that game trajectory similarity can be evaluated using a probabilistic interpretation of JSON-Bag with \textbf{Jensen-Shannon distance}, a probability distribution distance metric well-founded in information theory.
  \item We demonstrate the \textbf{sample efficiency} of JSON-Bag prototypes for game trajectory classes representation and their ability to \textbf{automatically extract game features} to be used in other methods of feature selection.
  \item We show that the JSD between JSON-Bag prototypes of different types of agents highly correlates with their behavioral difference measured by a policy distance.
\end{itemize}


\section{Background}
In traditional board game research, abundant work has been done to determine which game features to extract for different tasks (game-playing agents, game analysis, procedural content generation, etc.). In Go, Silver et. al. \cite{silver_temporal-difference_2012} use local board features in a value function approximator to build a strong playing agent. Browne \cite{browne_automatic_2008} introduces an extensive set of game metrics to evaluate the overall game structure and aesthetics in the context of automated board game generation. With the advances of deep learning, AlphaZero is a framework taking advantage of neural networks' ability for automatic feature learning and self-play reinforcement learning to build a superhuman Go, Chess, and Shogi agent without domain knowledge \cite{silver_mastering_2017}.

In video games, due to more complex game dynamics, hand-crafted features are difficult to define, so a common choice of game state representation is the raw pixels and the game memory values\cite{zhan_retrieving_2018}. Game sprite-sheets can also be used together with object recognition to learn a graph representation of the game system \cite{guzdial_automated_2018}. Player telemetry data is often used as game trajectory representation in playstyle analysis \cite{drachen_comparison_2013}.

Video game description languages (VGDL) are general frameworks of symbolic representation to describe game states and game systems. PyVGDL \cite{schaul_video_2013} and Ludii \cite{piette_ludii_2020} are examples of VGDL used for 2D arcade games and abstract board games, respectively. Although VGDLs are useful encodings for game generation and analysis, they are narrow in scope and not applicable to existing games without extensive manual effort.

Distance or similarity metrics are an essential component in game data mining to quantify the similarity between objects of interest, such as game trajectories in playstyle clustering \cite{drachen_comparison_2013}. These metrics are also used in Quality-Diversity (QD) algorithms to determine the novelty of new solutions, which have seen wide usage in game content generation \cite{liapis_transforming_2021} and game-playing agents \cite{jackson_novelty_2019}. The most common choices of metrics are the Euclidean distance and cosine similarity.

\section{Games}
The following games are used in our experiments, using their implementation in the TAG framework \cite{gaina_tag_2020}.

\textbf{7 Wonders} (Antoine Bauza, 2010): Players draft cards to build their civilizations, interacting with neighbors by passing cards and buying resources. Each player is randomly assigned a Wonder with unique special abilities. 

\textbf{Dominion} (Donald X. Vaccarino, 2008): Players build their deck by purchasing cards from a fixed common pool to create an ``engine" to acquire victory point cards in the late game. This paper uses the ``First Game" card set up. 

\textbf{Sea Salt and Paper} (Bruno Cathala, 2022): A set-collection game where players draw cards to create card combos that maximize their score. Players can choose when and how to end the round---higher risk options can gain or lose them points. 

We call the following games \textit{sparse}, meaning they have few unique game components. For example, \textit{7 Wonders} has coins, resources, and multiple unique cards, while \textit{Connect4} only has a grid with a unique board piece type per player.

\textbf{Connect4} (Milton Bradley, 1974): Players take turns dropping board pieces into a vertical grid to connect four of their pieces in a row, column, or diagonal before their opponent. $8 \times 8$ is the default grid size for this paper.

\textbf{Dots and boxes} (Edouard Lucas, 1889): Players take turns placing a link between adjacent dots on a grid, until filled. When a player forms a box on their turn, they score one point and play another turn.

\textbf{Can't Stop} (Sid Sackson, 1980): A push-your-luck dice game with eleven number tracks, from 2 to 12. Players roll four dice to form two sums and advance markers on corresponding number tracks, aiming to complete three tracks before their opponents. Players must decide whether to stop or keep rolling to advance and risk losing progress.

\section{JSON Bag-of-tokens model}
The JSON Bag-of-tokens model (JSON-Bag) represents game trajectories by tokenizing their JSON representation.
The JSON of a game state is made by serializing the data of every game component, with the specific serialization structure depending on the game implementation. The JSON of a game trajectory is simply an ordered-list of all the JSON game states.
\subsection{Tokenization}
We define a \textit{token} of a JSON as each of its individual \textit{atomic} components, identified by a string containing the path from the outermost level to that component, each level separated by a dot. In this paper, we define \textit{atomic} components as anything that is neither a dictionary nor a list. For example, a snippet from a \textit{7 Wonders} JSON:
\begin{lstlisting}[language=json,firstnumber=1]
{"currentAge": 2,
  "playerResources": [
    {"Wood": 2},
    {"Wood": 2}
]}
\end{lstlisting}
In this example, there are three atomic components on lines 1, 3, 4, all of which are integers. They can be tokenized as:
\begin{lstlisting}[language=json,numbers=none]
".currentAge.2",
".playerResources.Wood.2",
".playerResources.Wood.2"
\end{lstlisting}
Since \lstinline{"playerResources"} is a list storing the resources of each player, we may want to encode the ordering as well to retain information of player ownership:
\begin{lstlisting}[language=json,numbers=none]
".playerResources[0].Wood.2",
".playerResources[1].Wood.2"
\end{lstlisting}
These tokens are then processed into a JSON-Bag, following the same principles as "bag-of-words" from natural language processing \cite{berry_matrices_1999}, where the occurrences of each token are counted and combined into a token-occurrence count pair. Each game trajectory is a bag-of-tokens, and the occurrence values are normalized to sum to $1$ within a bag to model a probability distribution. For the rest of this paper, unless specified otherwise, JSON-Bag is normalized by default.

Using JSON-Bag does not entirely ignore the temporal aspects of game trajectories. E.g., in Poker, given the same final game state, having a specific card in hand at the beginning of the game would produce a JSON-Bag that is different from instead gaining that same card at a different time-step, since the former would have more occurrences of that card.

\subsection{Ordered vs. Unordered Tokenization}
A player's hand of cards, for example, can be stored internally as a list, but for many card games the order does not matter. In this case, using ordered-tokenization would make tokens that should have been the same into distinct tokens, causing loss of information. A more sophisticated tokenization procedure would be aware of which list to be tokenized with or without ordering.
In this paper, to demonstrate the simplicity of the method, we consider for each game to tokenize \textit{only} in-order, or \textit{only} unordered, or \textit{both}, meaning any token involving a list is processed with both ordered and unordered tokenization. \autoref{tab:tokenMode} shows which tokenization mode is used for each game, which is decided based on preliminary testing.

For all games, we did not make any fundamental changes to the data structure of the game object classes, only implementing the mechanism to serialize game states into JSON files. Our serialization does not include any explicit information on the actions taken by the players (e.g., action history).
\begin{table}[!t]
    \centering
    \begin{tabular}{|c|c|}
        \hline
        Game &  Tokenization mode\\
        \hline
        7 Wonders & unordered\\
        Dominion & unordered\\
        Sea Salt and Paper & unordered\\
        Can't Stop & both\\
        Connect4 & ordered\\
        Dots and boxes & ordered\\\hline    
    \end{tabular}
    \caption{Tokenization mode used for each game.}
    \label{tab:tokenMode}
\end{table}

\subsection{JSON-Bag Prototype}
A JSON-Bag prototype is simply the average of all bags of that specific class in the dataset, forming a single JSON-Bag that represents an entire class of game trajectories (e.g., all game trajectories generated by One-step-look-ahead (OSLA) agents). This is useful for cross-class analysis and prototype-based nearest-neighbor search (P-NNS).

\subsection{Interpretation}
\label{prob-model}
Traditionally, bag-of-words models in natural language processing (NLP) are often understood as vector space model \cite{berry_matrices_1999}, where a bag of token-frequency pairs is a point or a unit vector in the hyperplane when used with Euclidean distance or cosine similarity, respectively.

Instead, we interpret \textbf{JSON-Bag as probabilistic models of game trajectories}, meaning, the frequency value of a token is the probability that a game trajectory would ``generate" that token during gameplay. We argue that this is a more intuitive interpretation of JSON-Bag than the vector space model. In this interpretation, a class of game trajectories has a single true probability distribution for all the trajectories, and a JSON-Bag prototype is the maximum-likelihood estimation of such a distribution for that class. This allows usage of divergence and distance metrics such as Kullback–Leibler (KL) divergence \cite{kullback_information_1951} and Jensen-Shannon distance \cite{lin_divergence_1991} to analyze similarity between distributions.
Similar interpretations of bag-of-words models have seen success in both NLP \cite{storey_like_2020} and other areas such as biomedical time-series analysis \cite{wang_bag--words_2013}.

\section{Jensen-Shannon Distance}
Jensen-Shannon (JS) divergence \cite{lin_divergence_1991} is an information-theoretic divergence based on Shannon entropy to measure the similarity between distributions. JS divergence is symmetric and bounded between $[0, 1]$, with $1$ being identical distributions. The square-root of this gives Jensen-Shannon distance (JSD), which is a \textit{metric} \cite{endres_new_2003} (i.e., satisfies triangle inequality).

For discrete distributions $P$ and $Q$ defined on the same sample space $\mathcal{X}$, the KL divergence between $P$ and $Q$ is:
\begin{equation}
    \label{kl}
    D_{KL}(P||Q) = \sum_{x\in{\mathcal{X}}}P(x)\log{\frac{P(x)}{Q(x)}}
\end{equation}
Given mixture distribution $M = \frac{1}{2}(P+Q)$, the JS divergence between $P$ and $Q$ is:
\begin{equation}
    \label{jsdiv}
    D_{JS}(P||Q) = \frac{1}{2}D_{KL}(P||M) + \frac{1}{2}D_{KL}(Q||M)
\end{equation}
JS divergence can be thought of as a symmetrized and smoothed version of KL divergence, where instead of directly comparing $P$ and $Q$ and having to choose a reference distribution, $P$ and $Q$ are compared as the reference distributions against the average distribution $M$. The JSD between $P$ and $Q$ is then defined as:
\begin{equation}
    \label{jsd}
    Dist_{JS}(P, Q) = \sqrt{D_{JS}(P||Q)}
\end{equation}

To validate our approach, \autoref{pol-dist} compares the JSD between JSON-Bag prototypes and policy distance. Given two agents $A$ and $B$, we define the \textbf{policy distance} between $A$ and $B$ over a set of game states $\mathcal{S}$ as:
\begin{equation}
    \label{policy-dist}
    D_{\pi}(A, B) = \frac{1}{|\mathcal{S}|}\sum_{s\in{\mathcal{S}}}{Dist_{JS}(\pi_A(s), \pi_B(s))}
\end{equation}
where $\pi_A(s)$ is agent $A$ policy at game state $s$. In other words, the policy distance between two agents over $\mathcal{S}$ is the average JSD between their policies over all game states $s \in \mathcal{S}$.

\section{Experiments}
\subsection{Classifying game trajectories into classes of game agents, game parameters, and game seeds}
For each game, JSON-Bag model is tested on three classification tasks 
(two for \textit{Connect4} and \textit{Dots and boxes}),
using full game trajectories as data points:
\begin{itemize}
  \item \textbf{Game agents}: Five agents: Random, One-step-look-ahead (OSLA), and three variants of Monte Carlo Tree Search (MCTS) \cite{browne_survey_2012}. The OSLA agent adds a small random noise to each action value to break ties. For simultaneous action games, i.e. \textit{7 Wonders}, OSLA evaluates each action by doing a single random rollout until all players have acted. The first MCTS agent is a vanilla open-loop MCTS using a single set of parameters for every game with a budget of 64ms per decision (MCTS-V). Then, for each game, the other two players are tuned with N-Tuple Bandit Evolutionary Algorithm (NTBEA) \cite{lucas_n-tuple_2018} specifically for both game and budget the agent will run on: 64ms (MCTS64) and 128ms (MCTS128).
  \item \textbf{Game parameters}: Four sets of game parameters. Certain variables in the games are parameterized, such as grid size (\textit{Connect4}, \textit{Dots and boxes}), columns' sizes (\textit{Can't Stop}), cards' values (\textit{Sea Salt and Paper}), number of cards available (\textit{Dominion}), or resource price \textit{(7 Wonders}). Due to space, this is not a full list of parameters being varied, but certain games have larger parameter spaces than others. Each set of parameters is randomly generated within a predefined range for each parameter. 
  \item \textbf{Game seeds}: Four game seeds, each used to initialize the random number generator of the game instances.
\end{itemize}

For each class, 500 games are played by game agents to generate game trajectories. For the \textit{game agents} task, the agent of each class played against copies of itself to generate game trajectories. For \textit{game parameters} and \textit{game seeds}, all games are generated by copies of MCTS64 played at 32ms budget. All games are played with 4 players, except for \textit{Connect4} and \textit{Dots and Boxes}, which are played with 2. For all models, datasets are split 50/50 for training and testing\footnote{No validation set is needed since there is no hyperparameter tuning.}.

\textbf{Prototype-based Nearest Neighbor Search} (P-NNS) is a nearest neighbor classifier that classifies a data point into a class according to the closest prototype. With P-NNS as the classifier, JSON-Bag and JSD are compared against a baseline using handcrafted features and Euclidean distance in six games: \textit{7 Wonders}, \textit{Dominion}, \textit{Sea Salt and Paper}, \textit{Can't Stop}, \textit{Connect4}, \textit{Dots and boxes}. Similar to JSON-Bag prototypes, a prototype for hand-crafted features is a single feature vector averaging over all feature vectors of a class within the training data. Preliminary testing with K-nearest-neighbor (KNN) showed that KNN never outperformed P-NNS (for both JSON-Bag and baseline) while requiring more computations and parameter tuning.

\textbf{Random Forest} (RF) \cite{breiman_random_2001} is a simple, yet effective, predictive model for tabular data due to its implicit feature selection mechanism. A single decision tree (DT) chooses a feature to split at each level greedily based on how much the split improves model performance, which automatically excludes noisy and irrelevant features but can ignore useful features that are conditioned on other feature splits. Random Forests improve this by building smaller DTs using randomly sampled features and training data for each tree.

We treat individual token-frequency pairs in JSON-Bag as features and use RF for the above classification tasks.

\textbf{Hand-crafted features}: For every game, the features include game duration and scores at game end. Scores of a player are recorded periodically throughout a game trajectory into a score vector $\mathbf{s}$, a linear regression model is fitted to predict $s_i$: $w \times i + b = s_i$, where $i$ is the index of the score. We extract $w$ and $b$ for each player as features.
For individual games, we define a set of game-specific features to be aggregated from actions played and extracted from the final game state.

\textit{Dots and boxes} only uses the generic game features defined above. \textit{Connect4} does not use any score related features since it has no scoring, only win, lose, or draw at game end.

The number of hand-crafted features used for each game is detailed in \autoref{tab:feature_count}.
\begin{table}[!t]
    \centering
    \begin{tabular}{l c}
        \hline
        Game & No. features \\
        \hline
        7 Wonders & 47 \\
        Dominion & 37 \\
        Sea Salt and Paper & 46 \\
        Can't Stop & 17 \\
        Connect4 & 13 \\
        Dots and boxes & 8 \\
        \hline
    \end{tabular}
    \caption{Number of hand-crafted features for each game.}
    \label{tab:feature_count}
\end{table}
Full report on MCTS parameters, game parameters, and hand-crafted features is in the GitHub repo \footnote{https://github.com/dienn1/JSONBag}.

JSON-Bag is also evaluated with cosine similarity (JSON-Cosine) and Euclidean distance (JSON-L2). For these two methods, each token is instead normalized across bags so their minimum and maximum values are 0 and 1, respectively.

Additionally, a special version of JSON-Bag where \textit{individual characters} (e.g., 'I', 'f', '\}', etc.) of a JSON string is tokenized, called JSON-Char, is evaluated on all games.

\begin{table*}[t]
    \centering
    \resizebox{\textwidth}{!}{ 
    \begin{tabular}{l|c@{\hskip 4pt}c@{\hskip 4pt}c|c@{\hskip 4pt}c@{\hskip 4pt}c|c@{\hskip 4pt}c@{\hskip 4pt}c|c@{\hskip 4pt}c@{\hskip 4pt}c|c@{\hskip 4pt}c|c@{\hskip 4pt}c}
        \hline
        & \multicolumn{3}{c|}{7 Wonders} & \multicolumn{3}{c|}{Dominion} & \multicolumn{3}{c|}{Sea Salt and Paper} & \multicolumn{3}{c|}{Can't Stop} & \multicolumn{2}{c|}{Connect4} & \multicolumn{2}{c}{Dots and boxes} \\
        \hline
        Method & Agent & Param & Seed & Agent & Param & Seed & Agent & Param & Seed & Agent & Param & Seed & Agent & Param & Agent & Param \\
        \hline
        Hand-crafted & 0.696 & 0.476 & 0.573 & 0.911 & 0.996 & \textbf{0.462} & 0.425 & 0.854 & 0.856 & 0.474 & 0.414 & 0.504 & 0.476 & 0.943 & \textbf{0.702} & 0.852 \\
        JSON-Bag & 0.742 & \textbf{0.546} & 0.942 & \textbf{0.938} & \textbf{1.000} & 0.350 & \textbf{0.718} & 0.990 & \textbf{0.983} & \textbf{0.493} & \textbf{0.977} & \textbf{0.922} & \textbf{0.644} & \textbf{1.000} & 0.509 & \textbf{1.000} \\
        JSON-Char & 0.433 & 0.358 & 0.883 & 0.930 & \textbf{1.000} & 0.334 & 0.525 & 0.865 & 0.345 & 0.528 & 0.552 & 0.540 & 0.403 & 0.947 & 0.472 & 0.994 \\
        JSON-L2 & \textbf{0.758} & 0.525 & \textbf{0.980} & 0.831 & 0.992 & 0.369 & 0.587 & 0.974 & 0.963 & 0.467 & 0.613 & 0.827 & 0.522 & 0.954 & 0.660 & 1.000 \\
        JSON-Cosine & \textbf{0.756} & 0.528 & \textbf{0.980} & 0.873 & \textbf{1.000} & 0.396 & 0.665 & \textbf{0.995} & 0.977 & 0.469 & 0.712 & 0.915 & 0.600 & 0.982 & 0.655 & 1.000 \\
        \hline
    \end{tabular}
    }
    \vspace{1pt} 
    \centering
    \caption{P-NNS Classification Accuracy. Best performance in 95\% confidence interval is bolded.}
    \label{tab:pnns}

    \centering
    \resizebox{\textwidth}{!}{ 
    \begin{tabular}{l@{\hskip 22pt}|c@{\hskip 4pt}c@{\hskip 4pt}c|c@{\hskip 4pt}c@{\hskip 4pt}c|c@{\hskip 4pt}c@{\hskip 4pt}c|c@{\hskip 4pt}c@{\hskip 4pt}c|c@{\hskip 4pt}c|c@{\hskip 4pt}c}
        \hline
        & \multicolumn{3}{c|}{7 Wonders} & \multicolumn{3}{c|}{Dominion} & \multicolumn{3}{c|}{Sea Salt and Paper} & \multicolumn{3}{c|}{Can't Stop} & \multicolumn{2}{c|}{Connect4} & \multicolumn{2}{c}{Dots and boxes} \\
        \hline
        Method & Agent & Param & Seed & Agent & Param & Seed & Agent & Param & Seed & Agent & Param & Seed & Agent & Param & Agent & Param \\
        \hline
        3-Shot     & 0.604 & 0.296 & 0.545 & 0.906 & 1.000 & 0.262 & 0.582 & 0.973 & 0.874 & 0.435 & 0.710 & 0.710 & 0.530 & 1.000 & 0.280 & 0.999 \\
        5-Shot  & 0.631 & 0.327 & 0.628 & 0.919 & 1.000 & 0.275 & 0.612 & 0.985 & 0.934 & 0.457 & 0.793 & 0.779 & 0.562 & 1.000 & 0.304 & 1.000 \\
        10-Shot & 0.663 & 0.362 & 0.744 & 0.935 & 1.000 & 0.272 & 0.644 & 0.987 & 0.969 & 0.458 & 0.826 & 0.844 & 0.585 & 1.000 & 0.344 & 1.000 \\
        20-Shot & 0.691 & 0.401 & 0.839 & 0.936 & 1.000 & 0.299 & 0.676 & 0.990 & 0.975 & 0.474 & 0.903 & 0.897 & 0.611 & 1.000 & 0.390 & 1.000 \\
        40-Shot & 0.708 & 0.452 & 0.904 & 0.938 & 1.000 & 0.306 & 0.685 & 0.990 & 0.980 & 0.469 & 0.939 & 0.900 & 0.630 & 1.000 & 0.427 & 1.000 \\
        \hline
    \end{tabular}
    }
    \vspace{1pt} 
    \centering
    \caption{N-Shot classification accuracy with P-NNS using JSON-Bag and JSD.}
    \label{tab:n-shot}
    
    \centering
    \resizebox{\textwidth}{!}{ 
    \begin{tabular}{l|c@{\hskip 4pt}c@{\hskip 4pt}c|c@{\hskip 4pt}c@{\hskip 4pt}c|c@{\hskip 4pt}c@{\hskip 4pt}c|c@{\hskip 4pt}c@{\hskip 4pt}c|c@{\hskip 4pt}c|c@{\hskip 4pt}c}
        \hline
        & \multicolumn{3}{c|}{7 Wonders} & \multicolumn{3}{c|}{Dominion} & \multicolumn{3}{c|}{Sea Salt and Paper} & \multicolumn{3}{c|}{Can't Stop} & \multicolumn{2}{c|}{Connect4} & \multicolumn{2}{c}{Dots and boxes} \\
        \hline
        Method & Agent & Param & Seed & Agent & Param & Seed & Agent & Param & Seed & Agent & Param & Seed & Agent & Param & Agent & Param \\
        \hline
        Hand-crafted & 0.731 & 0.785\dag & 0.667 & 0.966 & 1.000 & 0.587 & 0.527 & 0.972 & 0.953 & 0.533 & 0.460 & 0.678 & 0.560 & 0.997 & \textbf{0.729} & \textbf{1.000} \\
        JSON-Bag & \textbf{0.785} & \textbf{0.841\dag} & \textbf{1.000} & \textbf{0.989} & 1.000 & \textbf{0.609\dag} & \textbf{0.726} & \textbf{0.992} & \textbf{0.995} & \textbf{0.548} & \textbf{0.999} & \textbf{0.983} & \textbf{0.868\dag} & \textbf{1.000} & 0.716\dag & \textbf{1.000} \\
        JSON-Char & 0.374 & 0.193 & \textbf{1.000} & 0.969 & 0.999 & 0.327 & 0.564 & 0.972 & 0.394 & 0.512 & 0.725 & 0.669 & 0.408 & \textbf{1.000} & 0.723\dag & 0.996 \\
        \hline
    \end{tabular}
    }
    \vspace{1pt} 
    \centering
    \caption{Random Forest Classification Accuracy. Best performance in 95\% confidence interval is bolded.\newline Accuracy increase of at least 0.2 from P-NNS is marked with \dag.}
    \label{tab:rf}
\end{table*}

\subsection{JSON-Bag prototype distance and policy distance}
In our experiment, $\mathcal{S}$ is generated with random play, the policy distance for each pair of agents is calculated by \autoref{policy-dist}. The policies for Random and OSLA agents at any state are estimated by repeatedly sampling an action from them $n$ times; we choose $n=100$. Due to computational cost, we instead estimate the policies of MCTS agents by using softmax over the visit counts of the root node's children.

We plot the JSD between JSON-Bag prototypes of agent classes against their policy distance (\autoref{fig:json-policy}) and calculate the Pearson coefficient between them (\autoref{tab:pearson_r}).

\subsection{Preventing data leakage}
In certain tasks, the class labels are explicitly encoded in game object data. For example, fixing a game seed for \textit{7 Wonders} also fixes the wonder board assignment of each player, so there is a direct mapping from game seed to the wonder board assignment. Therefore, for classifying \textit{7 Wonders} seed, any information related to the player wonder board is removed.
Similarly, in \textit{Connect4}, there is a variable in the grid board object explicitly storing its size. We do not serialize that variable for classifying \textit{Connect4} game parameters.

\section{Results and Discussion}
\subsection{Prototype-based Nearest Neighbor Search}
\begin{figure*}
    \centering
    \includegraphics[width=0.8\linewidth]{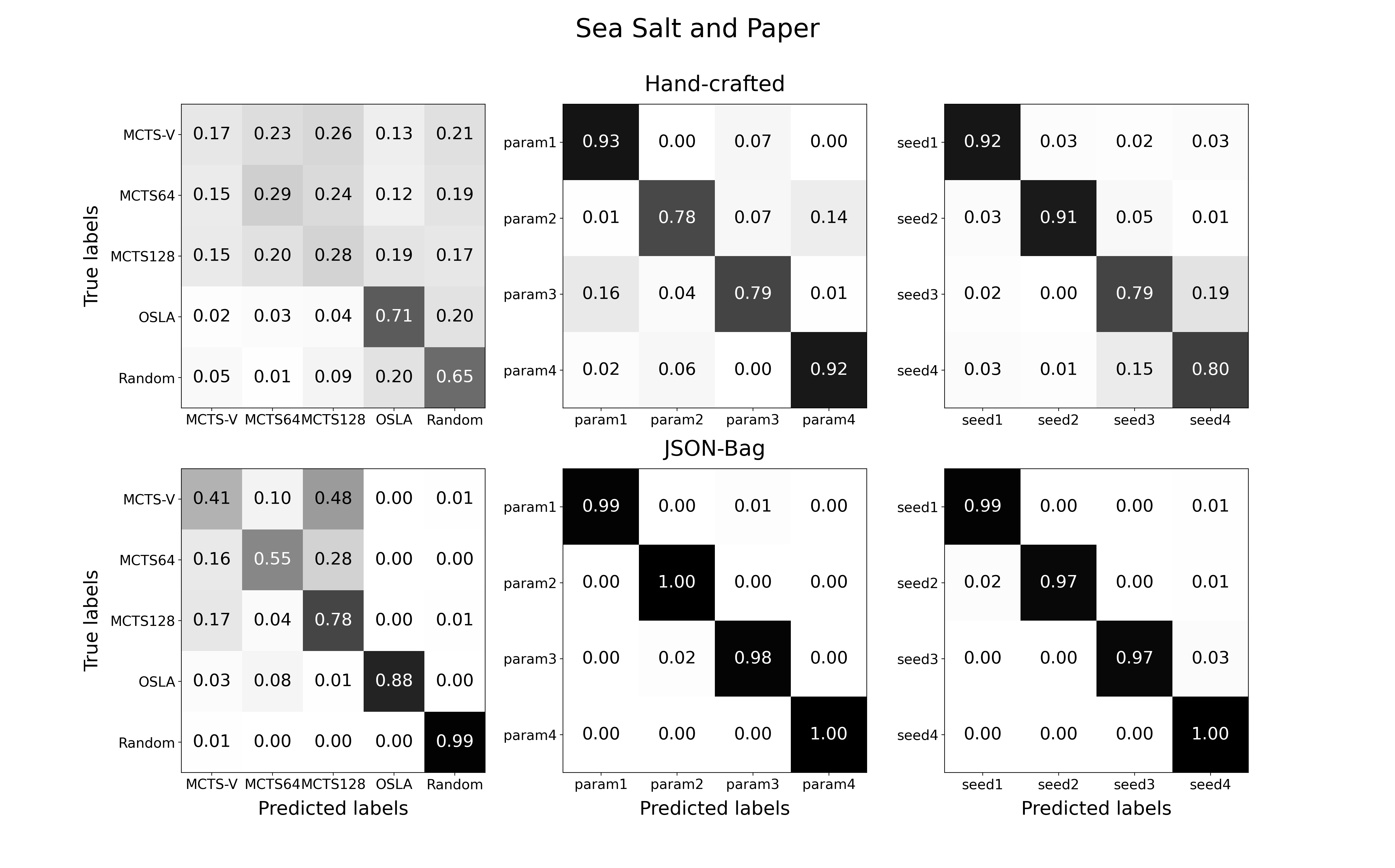}
    \caption{\textit{Sea Salt and Paper} Confusion Matrices with P-NNS. From left to right, classification of agents, game parameters, and game seeds. Top row shows results for hand-crafted features, bottom row for the JSON-Bag model. Darker shades in the diagonal represent higher classification accuracies.}
    \label{fig:ssp-pnns}
\end{figure*}

\begin{figure}
    \centering
    \includegraphics[width=0.75\linewidth]{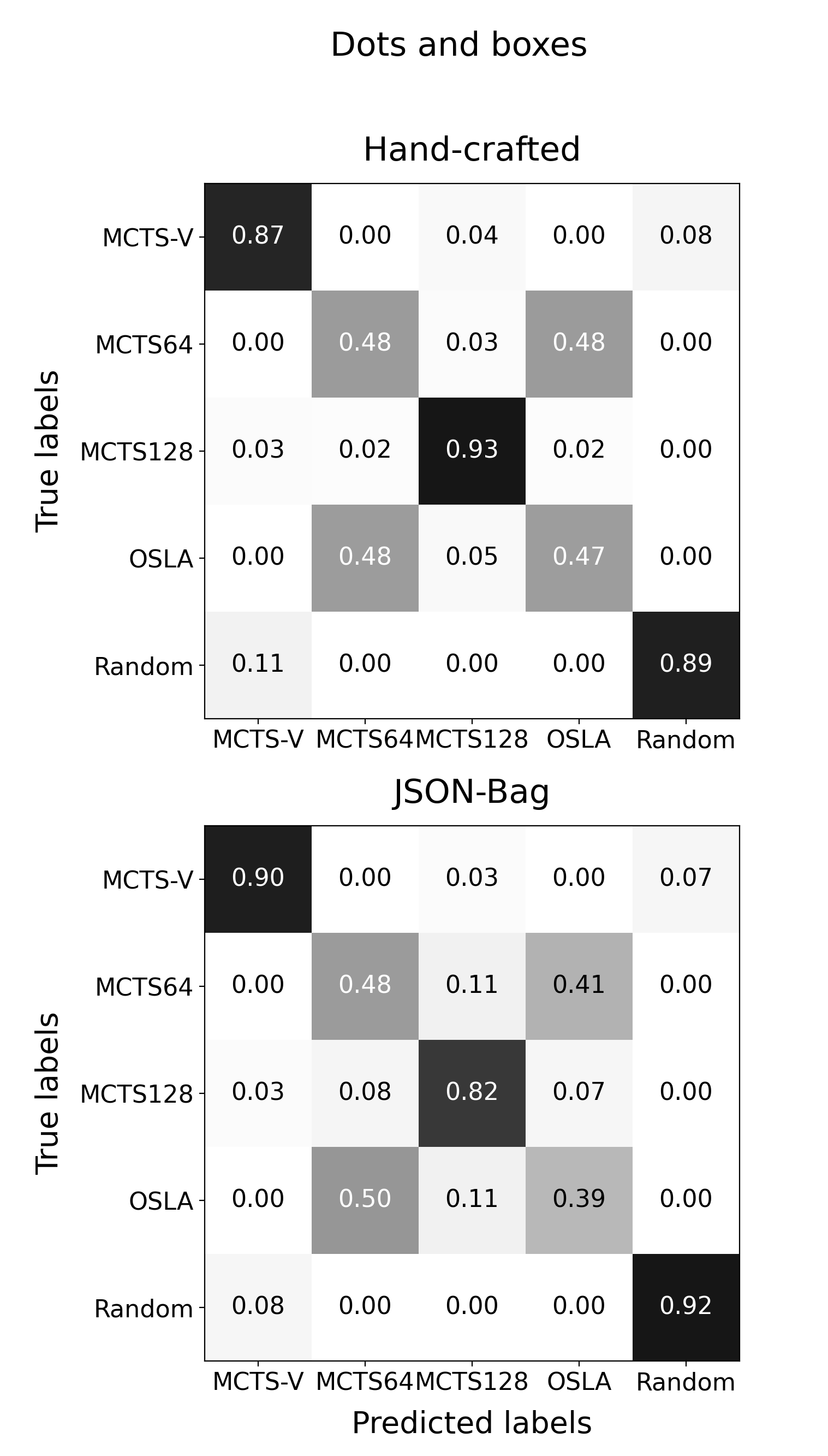}
    \caption{\textit{Dots and boxes} agents confusion matrices with RF.}
    \label{fig:dots-agent}
\end{figure}

\autoref{tab:pnns} shows the classification accuracy of P-NNS. JSON-Bag outperforms hand-crafted features on most tasks. Both JSON and hand-crafted approaches can mostly distinguish OSLA and Random agents from other MCTS agents, even if they struggle to differentiate between the MCTS agents. For example, in \textit{Sea Salt and Paper}, both approaches have difficulty separating the MCTS agents, but otherwise perform well on OSLA and Random (\autoref{fig:ssp-pnns}). Notably, JSON-Bag almost perfectly differentiates the MCTS agents from OSLA and Random; this same pattern is observed for most games, except for \textit{Connect4} and \textit{Dots and boxes}. For the former, it is possibly due to more complex interaction between tokens/features being required, since the pattern holds with RF; for the latter, our results suggest that MCTS64 is indistinguishable from OSLA in \textit{Dots and boxes} (see \autoref{fig:dots-agent}, \autoref{sec:dots}).

Comparing accuracy of JSD, cosine similarity, and Euclidean distance, JSD mostly outperforms the others, but their overall performance is comparable. We prefer JSD for the reasons mentioned in \autoref{prob-model}. 
Both methods perform poorly on classifying \textit{7 Wonders} game parameters and \textit{Dominion} game seeds. The former can be explained by small parameter space: we only parameterized the cost of buying resources from neighbors and the defined range of possible values is small, so different sets of game parameters may not behave differently enough for the models to distinguish them. The latter may be due to the role of randomness in \textit{Dominion}: the common card pool is fully-revealed and fixed for every game; the only stochastic element is in shuffling players' draw deck, whose contents are small and known to everyone (except for the ordering). This agrees with results from Goodman et al. \cite{goodman_seeding_2025} that game seeds have the least effect on \textit{Dominion}'s game outcome among all other tested stochastic games.

JSON-Bag significantly outperforms hand-crafted features in classifying parameters of \textit{Can't Stop}, game agents of \textit{Sea Salt and Paper}, and random seeds of \textit{7 Wonders} and \textit{Can't Stop}. This suggests JSON-Bag is extracting information that the hand-crafted features do not have.

On the other hand, using hand-crafted features significantly outperforms JSON-Bag in classifying \textit{Dots and boxes} agents with just 8 features. Using only turn count as a feature already reaches an accuracy of 61\%, compared to JSON-Bag's 51\%. The same information is also tokenized for JSON-Bag, but different turn counts are tokenized as completely different tokens, e.g. \lstinline{"turnCount.8"}, \lstinline{"turnCount.12"}. Instead, treating \lstinline{turnCount} as atomic and adding their values to the occurrence count may be more informative. This suggests that further refining of the tokenization method to treat specific types of atomic value differently would improve JSON-Bag.

It's worth emphasizing the accuracy of JSON-Char, where the \textit{individual characters} from JSON strings are tokenized. It is unexpected that JSON-Char would work at all, let alone achieve comparable accuracy or even outperform hand-crafted features in certain tasks. This demonstrates the potential of JSON format (or any data description format, especially human-readable ones) to be used as a medium for game analysis. The ``vocabulary" of JSON descriptions and its semantics are so limited and unambiguous (more so than in traditional NLP problems) that the frequency of individual characters is enough to characterise game trajectories. Applying more sophisticated NLP methods to JSON descriptions of game states is a promising direction for future research.

\subsection{Random Forest}
\label{forest}
\autoref{tab:rf} shows the classification accuracy of RF. JSON-Bag's accuracy on certain tasks where it underperforms with P-NNS significantly improves with RF, namely: \textit{7 Wonders} game parameters, \textit{Dominion} game seeds, \textit{Connect4} and \textit{Dots and boxes} playing agents. This big jump in accuracy implies JSON-Bag can be used together with more sophisticated machine learning methods capable of learning more complex interactions between its tokens.

If each token is considered a feature, then JSON-Bag automatically extracts features from game trajectories to be used as input for RF. This works particularly well with RF because of its implicit feature selection mechanism.

We believe hand-crafted features will outperform JSON-Bag with more effort into feature selection and engineering. In fact, JSON-Bag can help feature selection. For example, the Mean Decrease in Impurity (MDI) of a fitted RF as feature importance can inform which features from JSON-Bag are most relevant and should be manually considered. Detailed feature analysis using JSON-Bag is a topic for future work.

\subsection{N-Shot Classification}
We test N-shot classification with P-NNS, where the training data has N samples for each class. \autoref{tab:n-shot} shows P-NNS N-shot accuracy using JSON-Bag with JSD averaged over 20 trials. For ``easy" tasks (where regular P-NNS JSON-Bag has more than 80\% accuracy), N-shot P-NNS JSON-Bag can reach good accuracy with as few as 5 samples per class. This means P-NNS using JSON-Bag can efficiently differentiate classes that are ``obviously" different from each other. 

The sample efficiency is especially useful where two classes of game trajectories need their distance evaluated online (i.e., no trajectories have been generated prior). For example, when searching for playing agents with diverse playstyles, determining the novelty of new solutions requires many generated game trajectories to estimate game-specific metrics to be compared against existing agents \cite{guerrero-romero_map-elites_2021}. Instead, JSON-Bag and JSD would only need a few game trajectories to give a good estimate of how novel the new solution is. Its poor performance on more difficult tasks is arguably an acceptable downside, as it is likely those tasks involve classes that behave similarly. As shown in \autoref{fig:json-policy}, the policy distance between different agents is highly correlated to the JSD between their JSON-Bag prototypes. 

\subsection{JSON-Bag prototype distance and policy distance}
\label{pol-dist}
\begin{table}[!t]
    \centering
    \begin{tabular}{l c}
        \hline
        Game & Pearson-R \\
        \hline
        7 Wonders & 0.8824 \\
        Dominion & 0.773 \\
        Sea Salt and Paper & 0.6235 \\
        Can't Stop & 0.9688 \\
        Connect4 & 0.6225 \\
        Dots and boxes & 0.813 \\
        \hline
    \end{tabular}
    \centering
    \caption{Pearson coefficient correlation between JSON-Bag prototype distance and Policy distance between agent classes}
    \label{tab:pearson_r}
\end{table}
\begin{figure}
    \centering
    \includegraphics[width=0.8\linewidth]{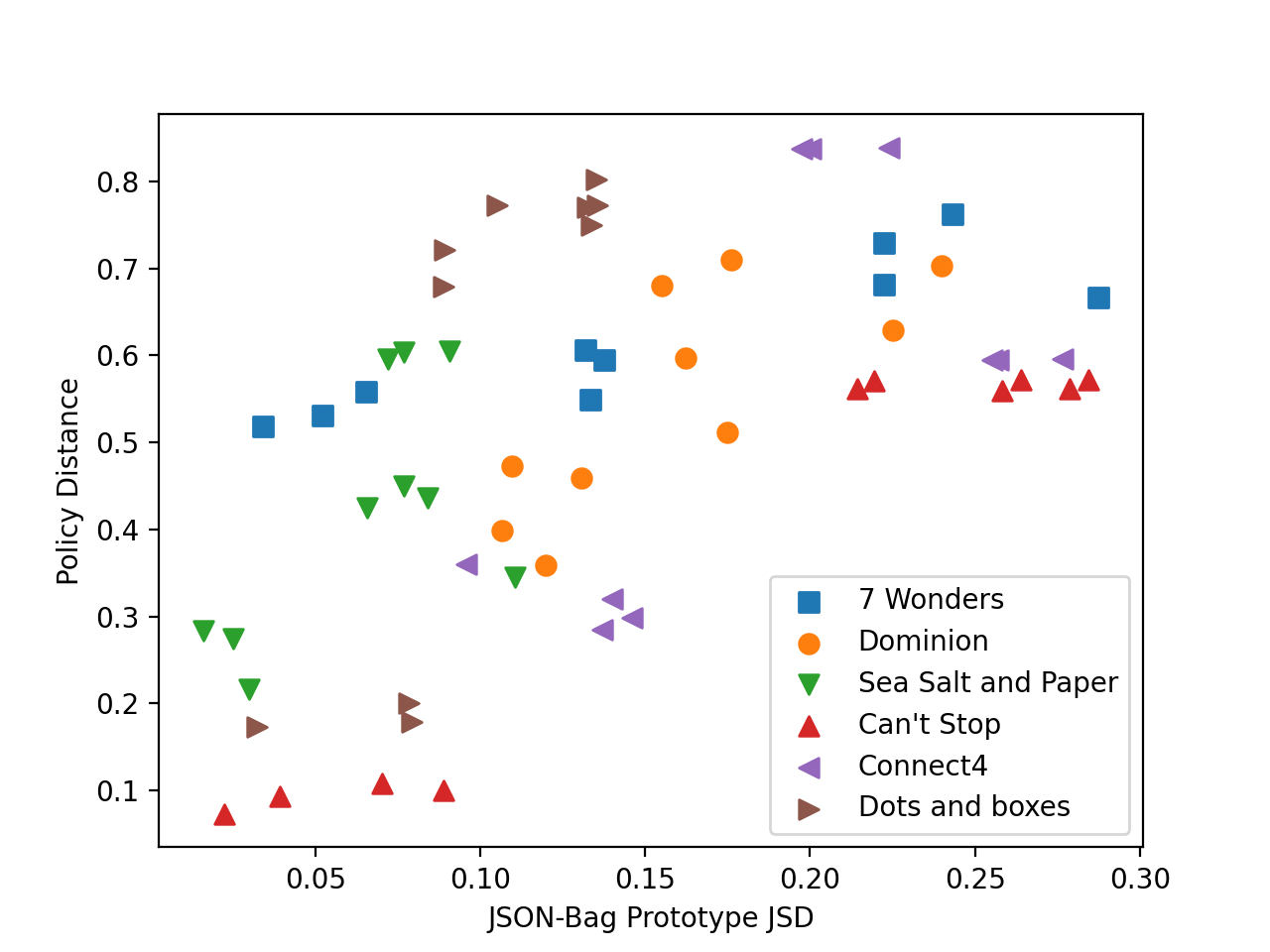}
    \caption{JSON-Bag prototype distance vs. Policy Distance between agent classes. Each point is a pair of agents.}
    \label{fig:json-policy}
\end{figure}
\autoref{fig:json-policy} plots the JSON-Bag prototype JSD against the policy distance between pairs of agents. \autoref{tab:pearson_r} detailed the Pearson correlation coefficient between them for each game, confirming that the two metrics are highly correlated. This validates JSON-Bag and JSD application in playstyle clustering without the need for hand-crafted features and game-specific metrics. Due to computational constraints, the states $\mathcal{S}$ are generated with random play. However, directly using policies of the two agents being evaluated instead may yield a policy distance that better reflects their actual behavioral difference.

\subsection{\textit{Dots and boxes} MCTS64 vs. OSLA}
\label{sec:dots}
\textit{Dots and boxes} tuned MCTS64 agent is an open-loop MCTS agent with maximum tree depth of 3, using progressive widening \cite{coulom_computing_2007} and MultiTree \cite{goodman_multitree_2022}.
However, it is completely indistinguishable from a simple OSLA agent for both JSON-Bag and hand-crafted features (\autoref{fig:dots-agent}). 
Looking at \textit{Dots and boxes} pairs of agents in \autoref{fig:json-policy}, we can see the pair with the lowest JSON-Bag Prototype JSD, MCTS64-OSLA, also corresponds to the lowest policy distance of all the pairs. This confirms that RF and P-NNS struggle to distinguish them because of similar behavior. A quick experiment of 200 games playing OSLA and MCTS64 against each other on \textit{Dots and boxes} showed that both achieve approximately 50\% win rate.

\subsection{Limitations}
As shown with \textit{Dots and Boxes}, JSON-Bag may not function well in ``sparse" games, where the information serialized from game state data is not as informative due to few unique game components, or games heavily dependent on spatial relations between components (e.g., grids). Further refinement to tokenization may improve this issue. For instance, exploring ways in which game components with a large range of values (e.g., turn count) could be represented to prevent all possible values from being considered as different tokens. For grid-based games, x and y coordinates of game elements can be tokenized in pairs instead (e.g., \lstinline{"x.6.y.9"} instead of individual \lstinline{"x.6"} and \lstinline{"y.9"}). 

\section{Conclusion}
We propose JSON Bag-of-Tokens model (JSON-Bag), a method to generically represent game trajectories by tokenizing the JSON descriptions of individual game states. We describe a domain-agnostic method of tokenizing JSON, with potential for further domain-specific refinement. A JSON-Bag is interpreted as a probabilistic model for game trajectories, which allows the use of Jensen-Shannon distance (JSD) metric to compare game trajectories. 
We evaluate the validity of our approach through various tasks.

First, six tabletop games are used as test bed---\textit{7 Wonders}, \textit{Dominion}, \textit{Sea Salt and Paper}, \textit{Can't Stop}, \textit{Connect4}, \textit{Dots and boxes}---each with three game trajectory classification tasks: classifying the playing agents, game parameters, or game seeds that generated the trajectories. Using prototype-based nearest neighbor search (P-NNS), JSON-Bag outperforms a baseline of hand-crafted features on the majority of tasks.

Second, we show that JSON-Bag prototype is a sample efficient representation for game trajectory classes through experiments with N-shot classification. This is especially useful when similarity between classes of game trajectories needs to be evaluated ``online". For example, when novelty of a new solution needs to be evaluated in novelty search, domain-specific metrics may need many simulation trajectories for reliable estimates \cite{guerrero-romero_map-elites_2021}, while JSON-Bag only needs a few.

Third, we demonstrate JSON-Bag's ability for automatic feature extraction by treating individual tokens-frequency pairs as features to use with Random Forest (RF), which significantly improves accuracy on tasks P-NNS underperform. Using JSON-Bag together with more sophisticated feature selection methods is an interesting direction of future research to aid in understanding game-specific features.

Finally, we show that the JSD between JSON-Bag prototypes of different agents highly correlates with behavioral difference measured by policy distance, validating JSON-Bag and JSD for playstyle clustering without hand-crafted features.

However, JSON-Bag can struggle with ``sparse" games, where serialized game state data is not as informative due to few unique game components, or games heavily dependent on spatial relations (e.g., grids), such as \textit{Dots and Boxes}. Further refinement to tokenization may help address these limitations.

Immediate future work should focus on testing JSON-Bag on more games, different JSON representations, and on more complex tasks to fully understand the limitations of this approach. The \textit{types} of games to be tested on should be extended beyond turn-based tabletop games.

Other NLP methods are also promising alternative approaches to JSON processing. We initially used Normalized Compression Distance (NCD) \cite{cilibrasi_clustering_2005} to measure the distance between raw JSON string of game trajectories. Despite recent success in topic modeling \cite{jiang_low-resource_2023}, NCD behaves inconsistently in the above tasks, and sometimes fails without a clear reason we can discern. Possible future work would be to analyze how to get NCD working and/or why it fails in this domain.


\section*{Acknowledgments}
For the purpose of open access, the author(s) has applied a Creative Commons Attribution (CC BY) license to any Accepted Manuscript version arising. This work was supported by the EPSRC Centre for Doctoral Training in Intelligent Games \& Games Intelligence (IGGI) EP/S022325/1.

\bibliographystyle{IEEEtran}
\bibliography{references}
\vspace{12pt}

\end{document}